\documentclass[10pt,twocolumn,letterpaper]{article}

\usepackage{cvpr}              
\usepackage{booktabs}
\usepackage{multirow}
\usepackage{tabularx}
\usepackage{makecell}
\usepackage{soul}
\usepackage[accsupp]{axessibility}
\newcolumntype{Y}{>{\raggedright\arraybackslash}X}
\definecolor{cvprblue}{rgb}{0.21,0.49,0.74}
\usepackage[pagebackref,breaklinks,colorlinks,allcolors=cvprblue]{hyperref}

\def\paperID{6} 
\def\confName{CVPR}
\def\confYear{2026}

\title{How Should Video LLMs Output Time? An Analysis of Efficient Temporal Grounding Paradigms}

\author{
Shengji Jin$^{1}$\thanks{Equal contribution.}\quad
Yuanhao Zou$^{1}$\footnotemark[1]\quad
Victor Zhu$^{2}$\quad
Zhengping Ji$^{2}$\quad
Chen Chen$^{1}$\\
$^{1}$The University of Central Florida \quad
$^{2}$Axon\\
{\tt\small j604319238@gmail.com, yuanhaoz@ucf.edu, chen.chen@crcv.ucf.edu}\\
{\tt\small vzhu@axon.com, zji@axon.com}
}

\begin{document}
\maketitle
\begin{abstract}
While Multimodal Large Language Models (MLLMs) have advanced Video Temporal Grounding (VTG), existing methods often couple output paradigms with different backbones, datasets, and training protocols. This makes it challenging to isolate the specific impact of the output design. Additionally, as VTG systems are increasingly considered for resource-constrained edge deployment, the trade-off between output formulation and system-level efficiency requires systematic investigation. In this paper, we present a controlled empirical study comparing three dominant VTG output paradigms: Text Numeral Generation, Temporal Token Generation, and Continuous Temporal Decoding. We evaluate these paradigms across identical compact VLMs (SmolVLM2, FastVLM, and Molmo2) using consistent datasets and LoRA fine-tuning protocols. Evaluations on Charades-STA, QVHighlights, and YouCook2 measure both localization accuracy and system efficiency, including inference latency, training throughput, and parameter overhead. Our results demonstrate that the choice of output formulation significantly affects both grounding accuracy and computational cost, independent of model scale. Specifically, the continuous distribution paradigm consistently achieves the most favorable efficiency-accuracy trade-off on the Pareto frontier, delivering robust localization with minimal latency overhead. These findings provide objective empirical guidelines for designing efficient, deployment-ready VTG systems. \footnote{Our code and models are publicly available: \url{https://tg-paradigms.github.io/}}
\end{abstract}    
\section{Introduction}
\label{sec:intro}

\begin{figure*}[t]
\centering
\includegraphics[width=\linewidth]{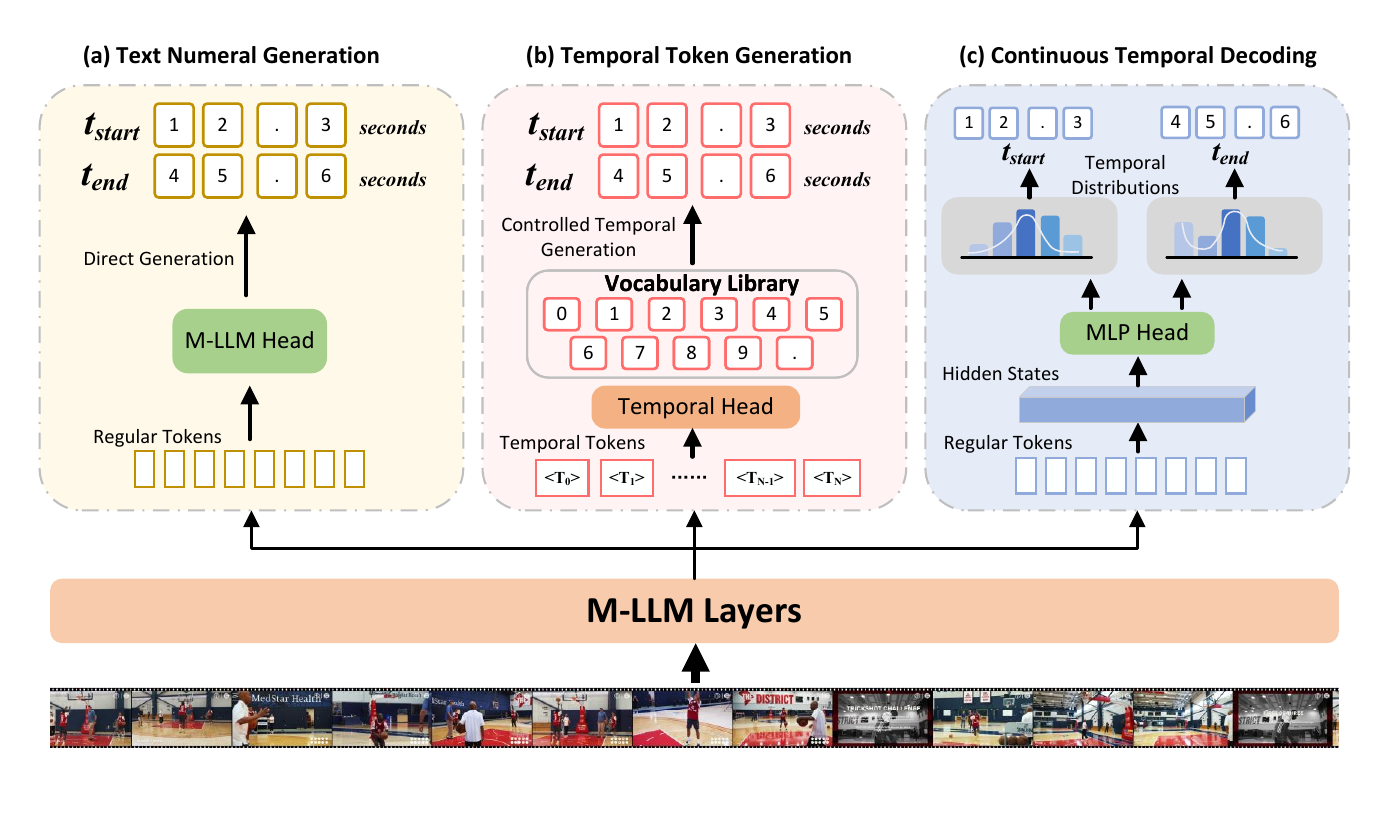}
\vspace{-1.5cm}
\caption{Overview of three output formulation paradigms for temporal grounding in Video LLMs. After extracting features through shared M-LLM Layers, the architecture diverges into three distinct designs: (a) \textbf{Text Numeral Generation} utilizes the standard M-LLM head to directly generate timestamps as sequences of regular text tokens. (b) \textbf{Temporal Token Generation} introduces specific temporal tokens and a dedicated temporal head, utilizing a restricted vocabulary library for controlled, sequential boundary generation. (c) \textbf{Continuous Temporal Decoding} maps hidden states through a lightweight MLP head to predict continuous temporal distributions, directly decoding boundary coordinates without auto-regressive generation.}
\label{fig:paradigms}
\vspace{-0.5cm}
\end{figure*}

Video temporal grounding (VTG), the task of localizing temporal segments in a video matching a natural language query, has attracted growing research interest. Core tasks including moment retrieval~\cite{gao2017tall, anne2017localizing}, dense video captioning~\cite{krishna2017dense, yang2023vid2seq}, highlight detection~\cite{lei2021detecting}, and grounded video question answering~\cite{xiao2024can, xiao2021next} all require models to produce precise temporal boundaries. As multimodal large language models (MLLMs) increasingly serve as the backbone for these tasks, diverse temporal output formulations have been proposed~\cite{wu2025survey}. Based on the different \textit{temporal output formulation}, existing approaches diverge into three main paradigms (Figure~\ref{fig:paradigms}): (a) text numeral generation, expressing timestamps as text numerals within the language modality~\cite{huang2024vtimellm, li2024groundinggpt}; (b) temporal token generation, introducing dedicated temporal tokens into the vocabulary~\cite{qian2024momentor, guo2024trace, guo2025vtg}; and (c) continuous temporal decoding, encompassing both regression modules that predict scalar timestamps from hidden states~\cite{moon2023correlation, wang2025internvideo2} and distribution-based methods that convert temporal embeddings into probability distributions over time bins~\cite{zeng2025distime}, both sharing the principle of mapping hidden representations to continuous temporal targets while differing in whether a distributional prior is imposed.

Each paradigm is exemplified by recent work: the text numeral approach generates temporal boundaries as plain-text numerals within a chat-style dialogue framework~\cite{huang2024vtimellm}; the temporal token generation paradigm employs causal event modeling, decoding events auto-regressively through task-specific encoder-decoder pairs~\cite{guo2024trace}; and continuous temporal decoding approaches, such as distribution-based methods, introduce a learnable temporal token to model boundary ambiguity via a probabilistic decoder~\cite{zeng2025distime}. We omit standalone regression heads from our study as they can be viewed as a degenerate case of distribution-based decoding without the distributional prior. However, as summarized in Table~\ref{tab:paradigm_compare_matrix}, each system uses its own backbone, data, and protocol, making it impossible to attribute performance differences to the output formulation alone.

This raises a critical yet underexplored question: \textit{how should temporal predictions be formulated in the model's output space?} The survey~\cite{wu2025survey} explicitly identifies this lack of controlled cross-paradigm comparison as an open challenge.

Moreover, these representative systems are all built on 7B-class or larger backbones. As the field moves toward practical deployment on resource-constrained platforms such as edge devices, mobile applications, and real-time video analytics, it becomes essential to understand how these output paradigms perform when paired with compact models, where computational budget is limited and every architectural choice carries amplified impact.

This paper addresses both gaps through a controlled empirical study. We deliberately choose compact backbones ranging from 0.5B to 8B parameters (SmolVLM2 0.5B/2.2B, FastVLM 1.5B, and Molmo2 4B/8B), covering a spectrum of resource-constrained deployment scenarios. We implement three representative temporal output paradigms under a unified framework, ensuring that backbone, raw data source, optimizer, epoch count, and fine-tuning protocol are identical across all conditions. Each paradigm naturally requires its own data formatting (e.g., structured event sequences for the generative paradigm, special token annotations for the distribution-based paradigm, and plain-text timestamp strings for the text numeral baseline), but the underlying video-query-annotation triples remain the same. The sole variable is thus the output formulation and its associated training objective. Our contributions are:

\begin{itemize}
    \item \textbf{Unified fine-tuning comparison.} We implement generative structured output (TRACE~\cite{guo2024trace}-style), distribution-based temporal modeling (DisTime~\cite{zeng2025distime}-style), and text numeral baseline (VTimeLLM~\cite{huang2024vtimellm}-style) paradigms on the same set of compact backbones, using identical training configurations to isolate the output formulation as the sole experimental variable.
    \item \textbf{Efficiency-aware analysis.} Beyond localization accuracy, we analyze parameter overhead, training stability, inference determinism, and computational cost, dimensions critical for practical deployment yet underexplored in the VTG-MLLM literature~\cite{wu2025survey}.
    \item \textbf{Empirical insights at compact scale.} We report how localization accuracy, training convergence, and inference determinism vary across paradigms when the backbone is scaled from 0.5B to 8B, and identify which output formulation offers the best accuracy-efficiency trade-off for deployment on resource-constrained platforms.
\end{itemize}

\begin{table*}[t]
\centering
\caption{Detailed comparison of existing VTG-MLLM systems. To highlight the extreme heterogeneity, we categorize the setups by LLM backbone, visual encoder, data scale, and optimization pipeline. \colorbox{blue!10}{1-S}, \colorbox{orange!10}{2-S}, and \colorbox{green!10}{3-S} denote 1-stage, 2-stage, and 3-stage training protocols, respectively.}
\vspace{-0.2cm}
\label{tab:paradigm_compare_matrix}
\scriptsize
\setlength{\tabcolsep}{4pt}
\renewcommand{\arraystretch}{1.2}

\begin{tabularx}{\textwidth}{@{}l l l l r c X@{}}
\toprule
\textbf{Paradigm} & \textbf{Method} & \textbf{LLM Backbone} & \textbf{Visual Encoder} & \textbf{Data Size} & \textbf{Stages} & \textbf{Architectural Additions / Key Designs} \\
\midrule

\multirow{6}{*}{\textbf{Text Numeral}}
& HawkEye~\cite{wang2024hawkeye} & Vicuna v1.5 (7B) & CLIP ViT-L/14 & 1.1M & \colorbox{green!10}{3-S} & Recursive search space narrowing \\
& SlowFocus~\cite{nie2024slowfocus} & Vicuna v1.5 (7B) & CLIP ViT-L/14 & 1.0M & \colorbox{green!10}{3-S} & Mixed-frequency sampling mechanism \\
& VTimeLLM~\cite{huang2024vtimellm} & Vicuna v1.5 (7B) & CLIP ViT-L/14 & 170K & \colorbox{green!10}{3-S} & None (Textual boundary format) \\
& TimeChat~\cite{ren2024timechat} & LLaMA-2 (7B) & EVA-CLIP ViT-G/14 & 125K & \colorbox{blue!10}{1-S} & Sliding video Q-Former \\
& GroundingGPT~\cite{li2024groundinggpt} & Vicuna v1.5 (7B) & CLIP ViT-L/14 & 1.4M & \colorbox{green!10}{3-S} & Modality adapters \\
& Chrono~\cite{meinardus2024chrono} & BLIP-2 (OPT-2.7B) & BLIP-2 ViT-g/14 & $\sim$12K & \colorbox{blue!10}{1-S} & None (Input sequence blueprint) \\
\midrule

\multirow{6}{*}{\textbf{Temporal Token}}
& Momentor~\cite{qian2024momentor} & LLaMA (7B) & CLIP ViT-L/14 & 10M & \colorbox{green!10}{3-S} & TPM (Anchor interpolation) \\
& VTG-LLM~\cite{guo2025vtg} & LLaMA-2 (7B) & EVA-CLIP ViT-G/14 & 217K & \colorbox{blue!10}{1-S} & Absolute time tokens \& slot compression \\
& LITA~\cite{huang2024lita} & Vicuna (7B) & CLIP ViT-L/14 & 500K & \colorbox{blue!10}{1-S} & Time tokens; SlowFast visual tokens \\
& TRACE~\cite{guo2024trace} & Mistral (7B) & CLIP ViT-L/14 & 2.8M & \colorbox{orange!10}{2-S} & Enc-Dec pairs (time/score/text) \\
& SeViLA~\cite{yu2023self} & Flan-T5 XL (3B) & ViT-G/14 (1B) & 125K+ & \colorbox{orange!10}{2-S} & Keyframe selection via Localizer-Answerer \\
& Grounded-VideoLLM~\cite{wang2024grounded} & Phi-3.5 / Vicuna & CLIP + InternVideo2 & 2.3M & \colorbox{green!10}{3-S} & Two-stream encoding with discrete tokens \\
\midrule

\multirow{4}{*}{\makecell[l]{\textbf{Continuous}\\\textbf{Temporal}}}
& InternVideo2.5~\cite{wang2025internvideo2} & InternLM2.5 (7B) & InternViT-6B & $\sim$10M & \colorbox{green!10}{3-S} & CG-DETR temporal regression head \\
& DisTime~\cite{zeng2025distime} & InternVL2.5 (8B) & InternViT-6B & 1.25M & \colorbox{blue!10}{1-S} & Distribution time Enc \& Dec (3-layer MLP) \\
& VideoMind~\cite{liu2025videomind} & Qwen2-VL (7B) & InternViT-6B & 480K & \colorbox{green!10}{3-S} & Timestamp decoder with Chain-of-LoRA \\
& TimeRefine~\cite{wang2026timerefine} & Vicuna v1.5 (7B) & CLIP ViT-L/14 & 1.0M & \colorbox{green!10}{3-S} & Auxiliary linear regression head + L1 offset \\

\bottomrule
\end{tabularx}
\vspace{-0.3cm}
\end{table*}
\section{Related Work}
\label{sec:related}

Traditional video temporal grounding methods employ task-specific architectures with large-scale video-text pretraining, such as Moment-DETR~\cite{lei2021detecting}, CG-DETR~\cite{moon2023correlation}, and UniVTG~\cite{lin2023univtg}. While effective, they are limited to single-task settings and lack zero-shot capability~\cite{wu2025survey}. MLLMs have since enabled unified, end-to-end temporal grounding across multiple tasks, but existing approaches diverge substantially in how temporal predictions are formulated in the output space~\cite{wu2025survey}. We organize related MLLM-based methods along this axis below.

\textbf{Text Numeral Generation.}
The most straightforward formulation represents timestamps as plain-text numerals within the language modality, reusing the LLM's native vocabulary without architectural modification. TimeChat~\cite{ren2024timechat} enhances temporal perception by conditioning a Q-Former on frame-level temporal descriptors to capture temporal context during feature extraction. VTimeLLM~\cite{huang2024vtimellm} adopts a multi-stage training pipeline that progressively aligns visual-textual modalities before fine-tuning on boundary-aware tasks. GroundingGPT~\cite{li2024groundinggpt} extends text numeral grounding across image, video, and audio modalities within a unified framework. Chrono~\cite{meinardus2024chrono} demonstrates that careful input-side temporal engineering, such as timestamp markers and video duration context, enables competitive grounding using only the standard LLM head with LoRA fine-tuning. While simple and parameter-free, this paradigm entangles temporal semantics with general number tokens, which may limit temporal precision~\cite{zeng2025distime}.

\textbf{Temporal Token Generation.}
To address the modal confusion inherent in text numeral approaches, a second line of work introduces dedicated temporal tokens into the vocabulary, creating a separate representational space for time. Momentor~\cite{qian2024momentor} injects temporal position encodings via learnable anchor points that define a continuous temporal space, enabling granular localization through interpolation. VTG-LLM~\cite{guo2025vtg} introduces zero-initialized absolute time embeddings to preserve the semantic integrity of pre-trained visual tokens. LITA~\cite{huang2024lita} adopts a relative time representation, segmenting videos into equal-length chunks and assigning unique temporal tokens to each segment. However, these token-based methods depend heavily on balanced time distributions during training and remain constrained by fixed-granularity binning~\cite{zeng2025distime}. TRACE~\cite{guo2024trace} takes a more structured approach via causal event modeling, employing a character-level vocabulary of 13 tokens and separate encoder-decoder pairs for timestamps, scores, and captions with an adaptive head-switching mechanism, demonstrating that task-specific structured decoding can yield strong multi-task VTG performance.

\textbf{Continuous Temporal Decoding.}
Rather than discretizing time into tokens, a third paradigm maps hidden-state representations directly to continuous temporal targets. The idea originates from CG-DETR~\cite{moon2023correlation}, which pioneered cross-modal attention mechanisms that predict temporal boundaries via regression heads. In the LLM era, InternVideo2.5~\cite{wang2025internvideo2} adapts this principle by attaching CG-DETR-style temporal perception heads to an MLLM backbone, predicting scalar timestamps from cross-modal hidden states. DisTime~\cite{zeng2025distime} moves beyond point regression by introducing a learnable $\langle\text{TIME\_STAMP}\rangle$ token whose hidden state is decoded through a lightweight MLP into probability distributions over temporal bins, explicitly capturing boundary ambiguity with minimal parameter overhead. Recent advancements further integrate continuous decoding into multi-step reasoning and iterative optimization frameworks. VideoMind~\cite{liu2025videomind} introduces an agentic workflow where a specialized \textit{Grounder} role utilizes a dedicated timestamp decoder, processing the hidden state of a $\langle\text{REG}\rangle$ token through a multi-layer transformer within a Chain-of-LoRA architecture. TimeRefine~\cite{wang2026timerefine} reformulates localization as a self-correcting process, employing an auxiliary regression head to predict iterative temporal offsets from an initial estimation. By supervising these residuals with an L1 loss, the model progressively sharpens temporal boundaries, bridging linguistic reasoning and continuous coordinate regression.

Despite the diversity of these approaches, each has been evaluated under distinct backbones, data, and protocols, leaving the relative merits of output formulations unclear~\cite{wu2025survey}.

\section{Paradigm Formulation}
\label{sec:paradigm}

We now formally define the three temporal grounding output paradigms. To establish a unified mathematical framework, let $I$ denote the textual instruction, $F$ represent the visual input, and $\mathbf{h} \in \mathbb{R}^{d}$ denote the semantic representation extracted by the VLM at a designated temporal query position. The objective is to predict the ground-truth temporal boundaries $t^* = (t_s^*, t_e^*)$. 

\subsection{Temporal Token Generation}

\textbf{General Formulation.} This paradigm abstracts temporal grounding as a structured sequence generation problem. Rather than treating time independently, the temporal boundaries $\mathcal{T}$ are decoded alongside associated metadata $\mathcal{M}$ (e.g., action categories, saliency scores) as a coherent event sequence $\mathcal{E} = \{e_1, e_2, \ldots, e_K\}$. The generation process is governed by a joint probability distribution, auto-regressively factorized over both the events and their internal structured attributes:
\begin{equation}
P(\mathcal{E} \mid I, F) = \prod_{k=1}^{K} P(e_k \mid e_{<k}, I, F),
\end{equation}
where $e_k = f(\mathcal{T}_k, \mathcal{M}_k)$ encapsulates the temporal boundaries and metadata, and $e_{<k}$ denotes the preceding generated events.
By explicitly disentangling temporal coordinates from natural language using dedicated tokenizers or heads, this paradigm aims to preserve the inherent structure of video events.

\noindent\textbf{Specific Instantiation.} In our controlled experiments, we instantiate this paradigm following the causal event modeling framework of TRACE~\cite{guo2024trace}. Specifically, each event is defined as a tuple $e_k = (t_k, s_k, c_k)$, containing timestamps, a saliency score, and a caption. The abstract factorization is concretely implemented as:
\begin{equation}
\begin{split}
P(e_k \mid e_{<k}, \cdot) = & \ P(t_k \mid e_{<k}, \cdot) \cdot P(s_k \mid t_k, e_{<k}, \cdot) \\
& \cdot P(c_k \mid s_k, t_k, e_{<k}, \cdot),
\end{split}
\end{equation}
where $\cdot$ represents the global conditions $I$ and $F$. To prevent interference with standard text generation, timestamps and scores are processed via an independent tokenizer with a specialized vocabulary (e.g., digits, $\langle \text{sep} \rangle$, $\langle \text{sync} \rangle$), optimized via task-specific cross-entropy losses.

\subsection{Continuous Temporal Decoding}

\textbf{General Formulation.} Instead of regressing deterministic coordinates or discrete tokens, this paradigm formulates temporal localization as a probability distribution estimation. For a given temporal query, the model predicts a probability density (or mass) function $p(\tau \mid \mathbf{h})$ over the continuous or discretized temporal space $\Gamma$. The final timestamp prediction $\hat{t}$ is typically decoded via the expectation over this temporal space:
\begin{equation}
\hat{t} = \mathbb{E}_{\tau \sim p(\tau \mid \mathbf{h})}[\tau] = \int_{\Gamma} \tau \cdot p(\tau \mid \mathbf{h}) d\tau,
\end{equation}
which naturally models prediction uncertainty and alleviates the ambiguity of subjective boundary annotations.

\noindent\textbf{Specific Instantiation.} We realize this paradigm by adopting the architecture from DisTime~\cite{zeng2025distime}. The abstract temporal space $\Gamma$ is discretized into $reg_{max} + 1$ bins. A dedicated token $\langle \text{TIME\_STAMP} \rangle$ is introduced to extract the representation $\mathbf{h}$. A Distribution-based Time Decoder maps $\mathbf{h}$ to discrete distributions $e_{st}$ and $e_{et}$ for start and end times via softmax, where $e_{st}^{(i)}$ and $e_{et}^{(i)}$ denote the probability mass at the $i$-th bin. The expectation integral is implemented as a weighted summation over discrete anchor points $a_i = i / reg_{max}$ to obtain the start and end predictions $\hat{t}_s$ and $\hat{t}_e$:
\begin{equation}
\hat{t}_s = \sum_{i=0}^{reg_{max}} e_{st}^{(i)} \cdot a_i, \quad \hat{t}_e = \sum_{i=0}^{reg_{max}} e_{et}^{(i)} \cdot a_i.
\end{equation}
The model is optimized using a combination of 1D-IoU regression loss and distribution focal loss (DFL).

\subsection{Text Numeral Generation}

\textbf{General Formulation.} As the most direct paradigm, this approach casts temporal grounding strictly as a standard natural language generation (NLG) task. The temporal boundaries $\mathcal{T}$ are serialized into a string of discrete text tokens $W = (w_1, w_2, \ldots, w_L)$, where each $w_i$ belongs to the native LLM vocabulary $\mathcal{V}_{\text{LLM}}$, and $w_{<j}$ denotes the previously generated text tokens. The decoding relies solely on the pre-trained causal language modeling head:
\begin{equation}
P(\mathcal{T} \mid I, F) = \prod_{j=1}^{L} P(w_j \mid w_{<j}, I, F).
\end{equation}
This paradigm requires no architectural modifications, completely depending on the LLM's intrinsic numerical reasoning capabilities.

\noindent\textbf{Specific Instantiation.} While the original VTimeLLM~\cite{huang2024vtimellm} formulates temporal boundaries as relative frame indices (e.g., from 00 to 99), introducing such temporal scaling would create a confounding variable in our controlled study. To ensure a strictly fair comparison across paradigms, we adapt this representation to directly generate absolute timestamps in seconds. The target boundaries are formatted into a unified natural language template (e.g., ``from 52.0 to 63.0 seconds''). The generation probability strictly follows the standard next-token prediction objective, unified with the overall text generation loss:
\begin{equation}
\mathcal{L}_{\text{text}} = -\sum_{j=1}^{L} \log P(w_j \mid w_{<j}, I, F).
\end{equation}
\section{Experiments}
\label{sec:experiments}

\subsection{Datasets}


\noindent\textbf{Training data.}
We compile a multi-source training set of approximately 1.2M temporally annotated samples covering $\sim$400K unique videos from 11 sources, spanning moment retrieval (66.8\%), grounded video question answering (21.5\%), and dense video captioning (11.6\%) tasks. The detailed source-level composition is provided in Table~\ref{tab:data} (Appendix). All paradigms share the same underlying video-query-annotation triples; each applies its own formatting pipeline to produce the required training format. Table~\ref{tab:format} in the Appendix illustrates how the same annotation is represented under each paradigm.

\noindent\textbf{Evaluation.}
We evaluate on three benchmarks. Charades-STA~\cite{gao2017tall} is a moment retrieval benchmark of short indoor activity videos ($\sim$30s average duration). QVHighlights~\cite{lei2021detecting} provides longer, diverse videos ($\sim$150s) annotated with both moment boundaries and frame-level saliency scores. YouCook2~\cite{zhou2018towards} is a dense video captioning benchmark of instructional cooking videos ($\sim$320s average duration) with temporally localized procedure steps. For moment retrieval, we report R1@IoU at thresholds 0.5 and 0.7, along with mIoU. For highlight detection on QVHighlights, we additionally report mAP and HIT@1. For dense video captioning on YouCook2, we report CIDEr, SODA\_c, and F1.

\begin{table*}[t]
\centering
\caption{Quantitative comparison of three temporal output paradigms across diverse video grounding benchmarks. Task abbreviations: \textit{MR} (Moment Retrieval), \textit{HD} (Highlight Detection), and \textit{DVC} (Dense Video Captioning). Both mAP and HIT@1 metrics on QVHighlights require native confidence or saliency score outputs for ranking. Because the \textit{Text} and \textit{Cont.} paradigms inherently lack such scoring mechanisms, these metrics are exclusively evaluated on the \textit{Gen.} paradigm via its dedicated ScoreTower. Variance across 3 random seeds for SmolVLM2-2.2B and Molmo2-4B is reported in Table~\ref{tab:variance} (Appendix).}

\label{tab:main}
\small
\setlength{\tabcolsep}{4pt}
\begin{tabular}{ll ccc ccccc ccc}
\toprule
 & & \multicolumn{3}{c}{Charades-STA} & \multicolumn{5}{c}{QVHighlights} & \multicolumn{3}{c}{YouCook2} \\
\cmidrule(lr){3-5} \cmidrule(lr){6-10} \cmidrule(lr){11-13}
 & & \multicolumn{3}{c}{MR} & \multicolumn{3}{c}{MR} & \multicolumn{2}{c}{HD} & \multicolumn{3}{c}{DVC} \\
\cmidrule(lr){3-5} \cmidrule(lr){6-8} \cmidrule(lr){9-10} \cmidrule(lr){11-13}
Backbone & Para. & R1@.5 & R1@.7 & mIoU & R1@.5 & R1@.7 & mIoU & mAP & HIT@1 & CIDEr & SODA\_c & F1 \\
\midrule
\multirow{3}{*}{SmolVLM2-0.5B}
 & Text  & 15.1  & 6.8  & 16.2  & 9.0  & 5.6  & 12.9  & --   & --  & 0.9  & 0.5  & 5.9  \\
 & Cont. & 41.8  & 22.5  & 43.4  & 46.9  & 31.0  & 50.4  & --   & --  & 1.4  & 1.0  & 9.0  \\
 & Gen.  & 20.1  & 9.5  & 21.9  & 14.7  & 9.6  & 18.6  & 20.0   & 21.8  & 1.2  & 0.9  & 7.9  \\
\midrule
\multirow{3}{*}{FastVLM-1.5B}
 & Text  & 19.5  & 8.9  & 19.9  & 11.7  & 7.6  & 16.5  & --   & --  & 1.5  & 0.8  & 7.5  \\
 & Cont. & 51.3  & 25.1  & 46.6  & 60.9  & 35.5  & 54.6  & --   & --  & 2.2  & 1.7  & 10.5  \\
 & Gen.  & 26.2  & 12.3  & 28.9  & 18.4  & 12.3  & 24.0  & 25.8   & 28.0  & 2.8  & 1.4  & 10.3  \\
\midrule
\multirow{3}{*}{SmolVLM2-2.2B}
 & Text  & 19.4  & 8.7  & 20.8  & 11.6  & 7.2  &  16.5  & --   & --  & 1.6  & 0.7  &  7.6 \\
 & Cont. & 50.8  & 24.3  & 46.6  & 60.1  & 34.6  & 54.4  & --   & --  & 2.5  & 1.8  & 11.5  \\
 & Gen.  & 25.8  & 12.2  & 28.1  & 18.9  & 12.3  & 23.9  & 25.6   & 27.9  & 2.6  & 1.3  & 11.4  \\
\midrule
\multirow{3}{*}{Molmo2-4B}
 & Text  & 22.3  & 10.0  & 23.9  & 13.3  & 8.3  & 19.0  & --   & --  & 1.9  & 1.0  & 8.7  \\
 & Cont. & 65.8  & 37.6  & 56.3  & 69.1  & 42.8  & 59.6  & --   & --  & 5.9  & 4.1  & 16.9  \\
 & Gen.  & 33.7  & 16.0  & 32.3  & 21.7  & 14.1  & 27.5  & 29.4   & 32.1  & 3.0  & 1.6  & 13.0  \\
\midrule
\multirow{3}{*}{Molmo2-8B}
 & Text  & 25.6  & 11.5  & 27.5  & 15.3  & 9.5  & 21.8  & --   & --  & 2.1  & 1.1  & 9.2  \\
 & Cont. & 66.8  & 39.3  & 57.1  & 72.6  & 46.1  & 61.2  & --   & --  & 6.1  & 4.4  & 18.2  \\
 & Gen.  & 37.1  & 19.1  & 35.1  & 24.9  & 16.2  & 31.5  & 33.8   & 36.8  & 3.1  & 1.6  & 13.2  \\
\bottomrule
\end{tabular}
\vspace{-0.4cm}
\end{table*}

\subsection{Implementation Details}

We adopt five compact VLM backbones: the SmolVLM2 series (0.5B and 2.2B scales, equipped with SigLIP vision encoders), FastVLM-1.5B (FastViTHD vision encoder), and Molmo2 series (4B and 8B scales, equipped with SigLIP-SO400M vision encoder). While existing VTG work typically relies on 7B-class models~\cite{guo2024trace,zeng2025distime,meinardus2024chrono}, our selection heavily emphasizes compact backbones ($\le$ 4B) suitable for deployment on resource-constrained platforms. The 8B model is deliberately included as a standard-scale control, allowing us to directly benchmark against existing paradigms and rigorously verify if the observed trends hold when scaling down to highly efficient regimes.

All three paradigms are implemented as modular additions to each backbone. We freeze the visual encoder and apply LoRA~\cite{hu2022lora} (rank 16, $\alpha$=32) to the language model. For the distribution-based paradigm, the Time Decoder and Time Encoder are 3-layer MLPs with $R$=32 temporal bins, trained end-to-end alongside the LoRA-adapted backbone. For the generative paradigm, separate TimeTower, ScoreTower, and SyncTower modules are added, each operating on a dedicated 13-token vocabulary. The text numeral paradigm reuses the LLM's native vocabulary and language modeling head without additional modules.


We train for 1 epoch using AdamW (learning rate 1e-4, cosine schedule, warmup ratio 0.03) with DeepSpeed ZeRO-2 in bfloat16 precision. Videos are uniformly sampled at 32 frames. To ensure a fair comparison, we fix visual encoder weights, LoRA configuration, data preprocessing, and hardware across all conditions. The only variable is the output head architecture and its associated loss function. We verify robustness by repeating training with 3 random seeds on SmolVLM2-2.2B and Molmo2-4B; standard deviations are $\leq$0.5 mIoU (Table~\ref{tab:variance}).

\noindent\textbf{Efficiency Metrics.}
Since the three paradigms differ solely in their output modules, we strictly isolate and quantify the computational overhead introduced by each paradigm beyond the shared LoRA-adapted backbone. For each configuration, we report: (1) \textit{additional trainable parameters} strictly introduced by the specific output head (excluding the shared LoRA parameters); (2) \textit{training throughput} (samples per second);  (3) \textit{inference latency} (milliseconds per query); and (4) \textit{peak GPU memory} usage. To ensure standardized benchmarking, both throughput and latency are profiled on a single NVIDIA H200 GPU with a batch size of 1. These metrics provide a comprehensive perspective on the Efficiency-Accuracy Pareto frontier, allowing practitioners to assess the real-world deployment costs of each output formulation.

\section{Results and Analysis}
\label{sec:results}

\subsection{Quantitative Comparison on Benchmarks}
\label{sec:main_results}

To evaluate localization capabilities across paradigms, we benchmark all combinations on three diverse datasets under strictly controlled protocols, reporting standard metrics in Table~\ref{tab:main}.

\noindent\textbf{The Dominance of Continuous Representation.} As observed in Table~\ref{tab:main}, the output formulation intrinsically dictates the upper bound of temporal grounding performance. Under a strictly controlled setting, the \textit{Continuous Temporal Decoding (Cont.)} paradigm consistently achieves the most robust localization accuracy across all tasks. When scaling to the Molmo2-8B backbone, the \textit{Cont.} approach yields an impressive mIoU of 57.1\% on Charades-STA and an R1@0.5 of 72.6\% on QVHighlights. In stark contrast, the \textit{Text Numeral (Text)} baseline at the exact same 8B scale plateaus at a mere 27.5\% mIoU. This performance gap strongly supports our hypothesis: \ul{treating continuous temporal boundaries as discrete text tokens introduces quantization errors and constrains the geometric localization capabilities of MLLMs}.

\noindent\textbf{Task-Specific Merits of Token Generation.} While the \textit{Temporal Token Generation (Gen.)} paradigm trails behind the continuous approach in strict boundary precision, its specialized architecture (e.g., ScoreTower) uniquely supports native frame-level saliency prediction. It is crucial to note that both the \textit{Text} and \textit{Cont.} paradigms inherently predict macroscopic temporal boundaries (spans) and lack the architectural mechanism to assign dense, frame-by-frame continuous saliency scores. Consequently, they cannot natively perform the strict Highlight Detection (HD) task on QVHighlights, resulting in missing HIT@1 evaluations (Table~\ref{tab:main}). The \textit{Gen.} paradigm, however, provides robust structured predictions for HD (e.g., HIT@1 of 36.8 on 8B), validating the necessity of dedicated architectural heads when fine-grained saliency ranking is explicitly required.

\begin{table}[t]
\centering
\vspace{-0.1cm}
\caption{Reference comparison with published zero-shot results on Charades-STA. $^\dagger$Our controlled setting on FastVLM-1.5B.}
\label{tab:performance_ranked}
\small
\setlength{\tabcolsep}{3.5pt} 
\begin{tabular}{llcc}
\toprule
Paradigm & Method (Backbone) & R1@.5 & mIoU \\
\midrule
\multirow{5}{*}{\makecell[l]{Text\\Numeral}} 
 & VTimeLLM (Vicuna-7B) & 27.5 & 31.2 \\
 & GroundingGPT (Vicuna-7B) & 29.6 & -- \\
 & TimeChat (LLaMA-2-7B) & 32.2 & -- \\
 & Chrono-GPT (GPT-4o) & 28.8 & 33.0 \\
 & HawkEye (Vicuna-7B) & 31.4 & 33.7 \\
\midrule
\multirow{5}{*}{\makecell[l]{Temporal\\Token}} 
 & Momentor (LLaMA-7B) & 26.6 & 28.5 \\
 & VTG-LLM (LLaMA-2-7B) & 33.8 & -- \\
 & Grounded-VideoLLM (Phi3.5) & 36.4 & 36.8 \\
 & TRACE (Mistral-7B) & 40.3 & 38.7 \\
 & SeViLA (Flan-T5 XL) & 15.0 & 18.3 \\
\midrule
\multirow{4}{*}{\makecell[l]{Continuous\\Temporal}} 
 & InternVideo2.5 (InternLM2.5-7B) & 43.3 & 41.7 \\
 & DisTime (InternVL2.5-8B) & 60.3 & 53.1 \\
 & VideoMind (Qwen2-VL-7B) & 59.1 & 50.2 \\
 & TimeRefine (Vicuna-7B) & 38.6 & 36.2 \\
\midrule
\multirow{3}{*}{\textbf{Ours$^\dagger$}} 
 & Text (FastVLM-1.5B) & 19.5 & 19.9 \\
 & Cont. (FastVLM-1.5B) & 51.3 & 46.6 \\
 & Gen. (FastVLM-1.5B) & 26.2 & 28.9 \\
\bottomrule
\end{tabular}
\vspace{-0.5cm}
\end{table}

\subsection{The Paradigm Dividend vs. Scaling Law}
\label{sec:scaling}

To contextualize our findings and investigate the interplay between output paradigms and scaling laws, we compare our controlled models against published SOTA baselines (Table~\ref{tab:performance_ranked}) and track their scaling trajectories (Figure~\ref{fig:scaling_pareto}a). 

These results reveal a counter-intuitive phenomenon in the Video-LLM domain. Our carefully controlled \textit{Continuous Temporal Decoding} instantiation on a mere 1.5B parameter backbone (R1@0.5 of 51.3\%) significantly outperforms multiple established 7B-class models utilizing alternative paradigms, such as TRACE (40.3\%) and Grounded-VidLLM (36.4\%).

This finding introduces a concept we term the \textbf{``Paradigm Dividend"}. As depicted in the scaling curves (Figure~\ref{fig:scaling_pareto}a), the paradigm gap remains structurally persistent from 0.5B up to 8B capacities. Strikingly, the \textit{Cont.} model at just 0.5B (mIoU of 43.4\%) outperforms the \textit{Text} paradigm scaled all the way up to 8B (mIoU of 27.5\%). The \textit{Text Numeral} approach exhibits a fundamentally flat scaling trajectory, unable to bridge the gap simply by relying on the enhanced reasoning capacity of larger LLMs. This demonstrates that optimizing the temporal output formulation provides an orthogonal, highly effective path to superior performance, offering a more efficient alternative to brute-force parameter scaling.

\subsection{Efficiency-Accuracy Pareto Trade-off}
\label{sec:efficiency}

\begin{figure*}[t]
\vspace{-0.2cm}
\centering
\includegraphics[width=0.95\linewidth]{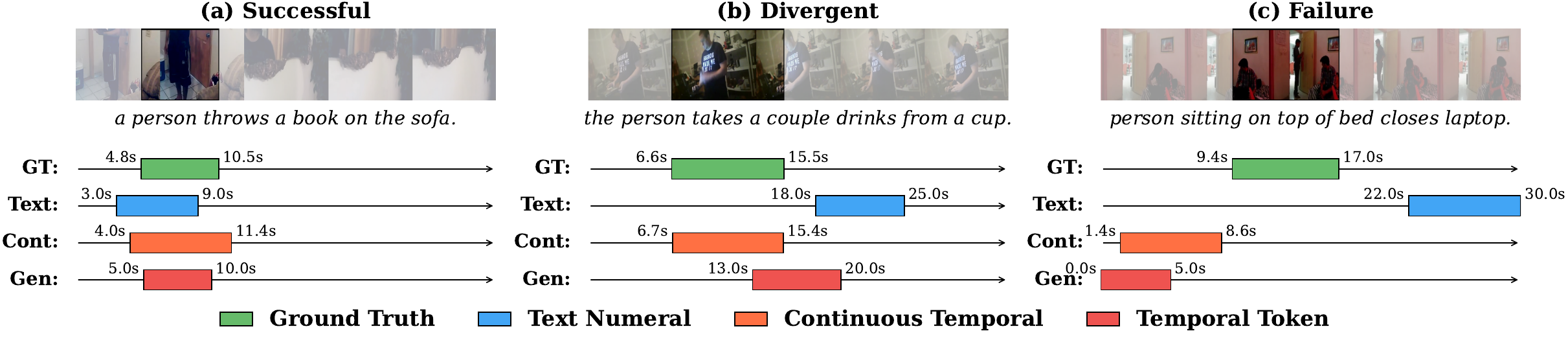}
\vspace{-0.3cm}
\caption{Qualitative comparison on Charades-STA (SmolVLM2-2.2B) across three difficulty levels. Each column shows sampled video frames and temporal predictions from all paradigms against the ground truth. (a)~Simple action; (b)~Temporally ambiguous query; (c)~Compound action leading to collective failure.}
\label{fig:qualitative}
\vspace{-0.5cm}
\end{figure*}

\begin{figure}[t]

\centering
\includegraphics[width=\linewidth]{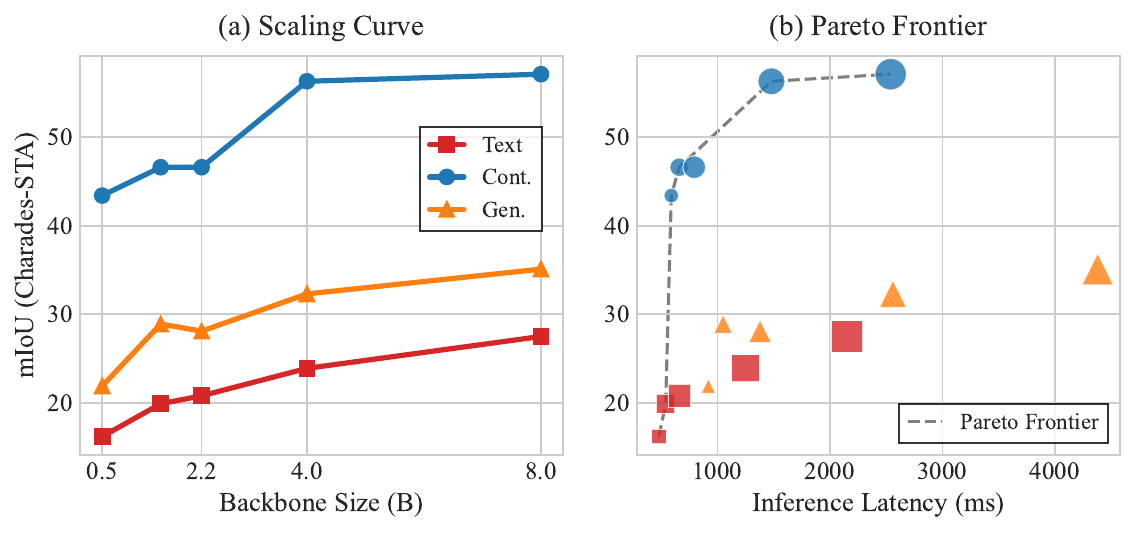}
\caption{(a)~Scaling behavior: mIoU vs.\ backbone size for each paradigm. (b)~Efficiency-accuracy Pareto frontier across all paradigm-backbone combinations. Both evaluated on Charades-STA.}
\label{fig:scaling_pareto}
\vspace{-0.5cm}
\end{figure}

For practical deployment on resource-constrained edge devices, it is critical to evaluate the system-level efficiency overhead introduced by each paradigm, rather than just their raw parameter counts. We visualize this via the efficiency-accuracy Pareto frontier in Figure~\ref{fig:scaling_pareto}(b). 

The \textit{Text Numeral} paradigm, despite introducing zero additional structural parameters, incurs noticeable inference delays due to the necessity to decode lengthy numerical strings auto-regressively. The \textit{Temporal Token (Gen.)} paradigm introduces minimal trainable parameters (111K) but incurs an extreme inference penalty (e.g., 1379 ms/query at 2.2B). This latency spike is directly attributed to the sequential, auto-regressive decoding of discrete temporal tokens interspersed within the semantic outputs. 

Notably, the \textit{Continuous Temporal (Cont.)} paradigm dominates the Pareto frontier. By leveraging a lightweight MLP decoder, it predicts boundary distributions in a \textit{single forward pass} over the localized hidden states. This non-autoregressive design sharply reduces inference latency (794 ms/query at 2.2B) compared to token generation, achieving the sweet spot between real-time responsiveness and rigorous localization precision. 

\noindent\textbf{Architectural Scaling Properties.} Table~\ref{tab:training_efficiency} also reveals a fundamental architectural divergence in how paradigm overheads scale with the LLM backbone. The \textit{Continuous Temporal (Cont.)} paradigm utilizes an MLP decoder, meaning its parameter count scales quadratically, $\mathcal{O}(D^2)$, with the LLM's hidden dimension $D$ (growing from 3.8M at 500M to 17.1M at 2.2B). In contrast, the \textit{Temporal Token (Gen.)} paradigm relies on linear embedding and projection heads (Towers), which scale linearly, $\mathcal{O}(D)$, leading to much smaller parameter footprints (52K at 500M vs. 111K at 2.2B). However, as demonstrated by our inference latency analysis (Figure~\ref{fig:scaling_pareto}b), despite the $\mathcal{O}(D^2)$ parameter growth of the \textit{Cont.} paradigm, its non-autoregressive forward pass completely bypasses the computational bottlenecks that plague the lightweight but sequential \textit{Gen.} paradigm.

\noindent\textbf{Synergy with Visual Compression.} An intriguing synergy emerges on the Pareto frontier when comparing FastVLM-1.5B and SmolVLM2-2.2B. Despite possessing 0.7B fewer parameters, the 1.5B \textit{Cont.} model matches the 2.2B model's localization accuracy (mIoU of 46.6\%) while significantly reducing inference latency (660 ms vs. 794 ms). This efficiency stems from the FastViT encoder's aggressive token downsampling, which couples exceptionally well with our non-autoregressive MLP decoder without degrading temporal boundaries. This finding highlights a critical system-level insight: \ul{co-optimizing input visual compression with a compatible continuous output formulation offers a more effective path to deployment efficiency than sheer LLM scaling}.
\subsection{Qualitative Analysis and Failure Cases}
\label{sec:qualitative}
\begin{table}[t]
\centering
\vspace{-0.1cm}
\caption{Training efficiency profile (single H200 GPU, batch size 1). \textit{Extra Params} accounts only for paradigm-specific heads.}
\label{tab:training_efficiency}
\resizebox{\linewidth}{!}{%
\begin{tabular}{@{}ll rcc@{}} 
\toprule
Backbone & Paradigm & Extra Params & Throu. (samp/s) $\uparrow$ & Mem. (GB) $\downarrow$ \\
\midrule
\multirow{3}{*}{SmolVLM2-0.5B}
 & Text  & 0 & 4.07 & 1.8 \\
 & Cont. & 3.8M & 4.09 & 1.8 \\
 & Gen.  & 52K & 4.12 & 1.9 \\
\midrule
\multirow{3}{*}{SmolVLM2-2.2B}
 & Text  & 0 & 4.03 & 5.5 \\
 & Cont. & 17.1M & 4.36 & 5.8 \\
 & Gen.  & 111K & 4.41 & 5.9 \\
\bottomrule
\end{tabular}%
}
\vspace{-0.5cm}
\end{table}
Figure~\ref{fig:qualitative} visualizes predictions on SmolVLM2-2.2B across three difficulty levels. 

For actions with clear visual cues (Figure~\ref{fig:qualitative}a), all paradigms successfully localize the ground truth. However, on temporally ambiguous queries (Figure~\ref{fig:qualitative}b), the \textit{Cont.} paradigm perfectly aligns with the boundaries (IoU of 0.99), as its continuous probability distribution gracefully models temporal uncertainty. Conversely, the \textit{Text} approach yields a severely bloated window (IoU of 0.66), and the \textit{Gen.} paradigm completely fails by latching onto irrelevant segments (IoU of 0.00).

Importantly, all paradigms fail on complex, multi-step actions requiring causal reasoning (Figure~\ref{fig:qualitative}c, ``\textit{person... closes laptop}"). The models mistakenly localize disjointed sub-actions or static background elements rather than the complete temporal logic. This reveals a critical limitation: while optimizing the \textit{output} paradigm maximizes boundary precision and efficiency, resolving deep temporal causality ultimately demands fundamental innovations in \textit{input-side} video encoding.

\subsection{Ablation Study}
\label{sec:ablation}

\begin{figure}[!t]
\centering
\vspace{-0.2cm}
\includegraphics[width=\linewidth]{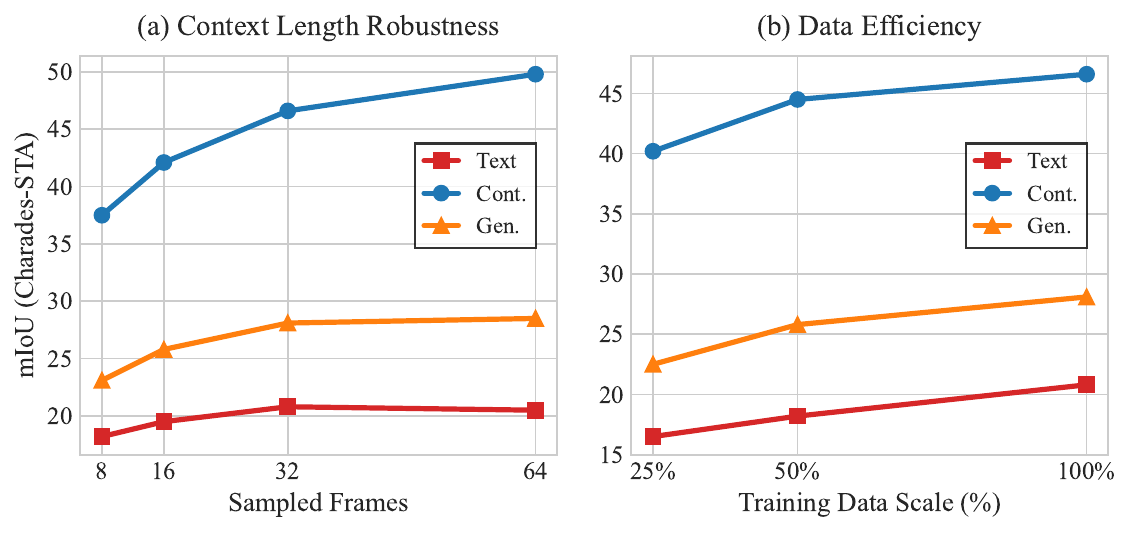}
\caption{Ablation studies evaluated on Charades-STA. (a) Context length robustness evaluated with varying sampled video frames (from 8 to 64). (b) Data efficiency evaluated under varying percentages of the training corpus (25\% to 100\%). Both studies use the SmolVLM2-2.2B backbone.}
\label{fig:ablation}
\vspace{-0.5cm}
\end{figure}

\noindent\textbf{Context Length Robustness.} Figure~\ref{fig:ablation}(a) shows scaling behavior from 8 to 64 frames. While all paradigms improve initially, the \textit{Text} paradigm saturates early and slightly degrades at 64 frames (mIoU of 20.5\%). This indicates that autoregressively aligning discrete numerals with elongated visual sequences leads to severe "context dilution." In contrast, the \textit{Cont.} paradigm efficiently digests dense visual cues, scaling steeply to 49.8 mIoU. Its non-autoregressive MLP decoder proves immune to sequence dilution.

\noindent\textbf{Data Efficiency.} Figure~\ref{fig:ablation}(b) highlights the \textit{Cont.} paradigm's superior data efficiency. Fine-tuned on merely 25\% of the training corpus, it achieves 40.2\% mIoU, substantially outperforming the \textit{Text} (20.8\%) and \textit{Gen.} (28.1\%) paradigms trained on 100\% of the data. Regressing a continuous temporal distribution is intrinsically more congruent with the localization task than coercing LLMs to memorize numeral-to-time mappings, thus significantly reducing the reliance on large-scale temporal annotations.

\noindent\textbf{Invariance to Visual Compression.} Beyond temporal length, we investigate spatial compression by contrasting SmolVLM2-2.2B ($\sim$81 visual tokens/frame) and FastVLM-1.5B ($\sim$64 tokens/frame). The \textit{Text} paradigm, reliant on fine-grained spatial details, drops from 20.8\% to 19.9\% mIoU under FastVLM's aggressive compression. Remarkably, the \textit{Cont.} paradigm maintains an identical 46.6\% mIoU across both backbones. This confirms that macroscopic distribution regression provides immunity to spatial detail loss.
\section{Conclusion}
\label{sec:conclusion}
We present a controlled empirical study comparing three output paradigms for MLLM-based video grounding: text numeral, temporal token, and continuous decoding. By unifying visual encoders, training data, and optimization protocols across compact backbones (0.5B to 8B), we isolate the output formulation as the sole experimental variable. Results show continuous decoding consistently achieves superior localization accuracy, highest throughput, and the optimal efficiency-accuracy trade-off. Crucially, optimizing this formulation unlocks a \textbf{``paradigm dividend''}, allowing compact sub-2B models to significantly outperform established 7B-class baselines and paving an efficient path for resource-constrained edge deployment.

While our findings establish the output formulation as a critical design axis, several open questions remain. First, our study fixes the input encoding; investigating how input-side temporal representations, such as learnable position embeddings or interleaved timestamp tokens, interact with each output paradigm is a promising direction. Second, the \textit{Continuous} paradigm currently lacks native saliency scoring, limiting its applicability to highlight detection. Augmenting the MLP decoder with a lightweight saliency branch that predicts per-frame scores alongside boundary distributions in a single forward pass could unify moment retrieval and highlight detection without sacrificing non-autoregressive efficiency.
{
    \newpage
    \small
    \bibliographystyle{ieeenat_fullname}
    \bibliography{main}
}
\newpage
\appendix
\setcounter{table}{0}
\renewcommand{\thetable}{S\arabic{table}}
\setcounter{figure}{0}
\renewcommand{\thefigure}{S\arabic{figure}}
\setcounter{equation}{0}
\renewcommand{\theequation}{S\arabic{equation}}
\section*{Appendix}
\section*{Overview}
\label{app:overview}
This supplementary material provides additional details and comprehensive benchmarks omitted from the main text due to space constraints. The contents are organized as follows:
\begin{itemize}
    \item \textbf{Section~\ref{app:implementation}:} Details our training data composition, hyper-parameters, data formatting templates, exact inference prompt designs, and statistical variance analysis across multiple random seeds.
    \item \textbf{Section~\ref{app:survey_table}:} Presents a macro-level survey and zero-shot performance leaderboard of existing VTG-MLLMs.
    \item \textbf{Section~\ref{app:qualitative}:} Provides extended qualitative visualizations to further illustrate the intrinsic behaviors of different output paradigms.
\end{itemize}
\noindent Upon acceptance, we will publicly release the full training and evaluation codebase, including all paradigm implementations, data formatting pipelines, and benchmark configurations.
\section{Additional Implementation Details}
\label{app:implementation}
\noindent\textbf{Hardware and Optimization.} Training is conducted on 2 nodes, each equipped with 4 NVIDIA H200 GPUs. We use a per-GPU batch size of 4 and a gradient accumulation step of 4, yielding a global effective batch size of 128. For LoRA fine-tuning, we set the rank $r=16$ and $\alpha=32$, targeting the $q, k, v, o$ projections in the LLM attention blocks. The learning rate is initialized at $1\times 10^{-4}$ with a cosine decay schedule and a 3\% linear warmup ratio.

\noindent\textbf{Training Data Composition.} Table~\ref{tab:data} details the source-level composition of our training set. The combined corpus comprises approximately 1.2M temporally annotated samples from $\sim$400K unique videos across 11 sources, spanning moment retrieval (66.8\%), grounded video question answering (21.5\%), and dense video captioning (11.6\%) tasks.

\begin{table}[h]
\centering
\caption{Training data composition by source.}
\label{tab:data}
\small
\begin{tabular}{lrr}
\toprule
Source & Samples & Videos \\
\midrule
InternVid~\cite{wang2023internvid}       & 608K & 87K  \\
YTTemporal~\cite{zellers2021merlot}      & 278K & 21K  \\
Valley~\cite{luo2023valley}              & 229K & 229K \\
DiDeMo~\cite{anne2017localizing}          & 33K  & 8K   \\
ShareGPT4Video~\cite{chen2024sharegpt4video}  & 24K  & 24K  \\
ViTT~\cite{huang2020multimodal}          & 8K   & 5K   \\
TextVR~\cite{wu2025large}                & 8K   & 8K   \\
COIN~\cite{tang2019coin}                 & 8K   & 8K   \\
ActivityNet~\cite{caba2015activitynet}    & 7K   & 7K   \\
QueryD~\cite{oncescu2021queryd}           & 5K   & 1K   \\
VideoChat~\cite{li2025videochat}          & 2K   & 2K   \\
\midrule
\textbf{Total}  & \textbf{$\sim$1.2M} & \textbf{$\sim$400K} \\
\bottomrule
\end{tabular}
\end{table}

\noindent\textbf{Data Formatting Setup.} To strictly isolate the output formulation as the sole experimental variable, all paradigms share the exact same raw video-text pairs but undergo paradigm-specific prompt and target formatting. We detail these exact input-output formats for a specific training sample in Table~\ref{tab:format}.

\begin{table}[h]
\centering
\caption{Backbone-level specifications.}
\label{tab:backbone_specs}
\small
\resizebox{\columnwidth}{!}{%
\begin{tabular}{@{}lllr@{}}
\toprule
Backbone & Vision Enc. & LLM & $d$ \\
\midrule
SmolVLM2-0.5B & SigLIP-B/16    & SmolLM2-360M &  960 \\
SmolVLM2-2.2B & SigLIP2-SO400M & SmolLM2-1.7B & 2048 \\
FastVLM-1.5B  & FastViTHD      & Qwen2-1.5B   & 1536 \\
Molmo2-4B     & SigLIP2-SO400M & Qwen3-4B     & 2560 \\
Molmo2-8B     & SigLIP2-SO400M & Qwen3-8B     & 4096 \\
\bottomrule
\end{tabular}%
}
\end{table}

\begin{table*}[h]
\centering
\small
\renewcommand{\arraystretch}{1.1}
\resizebox{\textwidth}{!}{%
\begin{tabular}{@{}p{\textwidth}@{}}
\toprule
\textbf{Shared annotation:} \texttt{video: split\_video\_7w/RP6nWBaB2K8.mp4} \quad ground truth: $[52.0,\, 63.0]$\,s \\
\midrule
\\[-6pt]
\textbf{(a) Text Numeral Generation} \\[2pt]
\begin{tabular}{@{\hspace{6pt}}r@{\hspace{4pt}}p{0.88\textwidth}}
\textsc{Input:} & \texttt{<video>{\textbackslash}n}With the input: `The worker cleans up excess plaster from the ground, ensuring a tidy workspace.', identify the precise second in the video where this event occurs. \\
\textsc{Output:} & The event happens from 52.0 to 63.0 seconds. \\
\textsc{Supervision:} & standard next-token cross-entropy (no additional temporal modules) \\
\end{tabular} \\[6pt]
\midrule
\\[-6pt]
\textbf{(b) Temporal Token Generation} \\[2pt]
\begin{tabular}{@{\hspace{6pt}}r@{\hspace{4pt}}p{0.88\textwidth}}
\textsc{Input:} & \texttt{<video>{\textbackslash}n}With the input: `The worker cleans up excess plaster from the ground, ensuring a tidy workspace.', identify the precise second in the video where this event occurs. \\
\textsc{Output:} & \texttt{<sync>}$\underbrace{\texttt{<time>}\,\cdots\,\texttt{<time>}}_{14}$\texttt{<score>}The worker cleans up excess plaster from the ground, ensuring a tidy workspace. \\
\textsc{Supervision:} & \texttt{times: [[52.0, 63.0]], scores: [[]]} $\;\rightarrow\;$ dedicated 13-token vocabulary \\
\end{tabular} \\[6pt]
\midrule
\\[-6pt]
\textbf{(c) Continuous Temporal Decoding} \\[2pt]
\begin{tabular}{@{\hspace{6pt}}r@{\hspace{4pt}}p{0.88\textwidth}}
\textsc{Input:} & \texttt{<video>{\textbackslash}n}With the input: `The worker cleans up excess plaster from the ground, ensuring a tidy workspace.', identify the precise second in the video where this event occurs. \\
\textsc{Output:} & The event happens in \texttt{<TIME\_STAMP>}. \\
\textsc{Supervision:} & \texttt{times: [[52.0, 63.0]]} $\;\rightarrow\;$ distribution-based time decoder \\
\end{tabular} \\[4pt]
\bottomrule
\end{tabular}%
}
\caption{Training data format comparison across the three paradigms for the same moment retrieval annotation. All paradigms share identical video inputs and queries; only the output format and temporal supervision differ.}
\label{tab:format}
\end{table*}

\noindent\textbf{Inference Prompt Templates.} In the era of Multimodal Large Language Models, the formulation of textual prompts profoundly impacts evaluation performance. To ensure rigorous reproducibility of our empirical study, we provide the exact inference prompt templates used across the three video temporal grounding tasks. During evaluation, the placeholder \texttt{\{query\}} is dynamically replaced with the specific textual query from the corresponding dataset. The detailed templates are presented in Table~\ref{tab:prompts}.

\begin{table*}[h]
\centering
\small
\renewcommand{\arraystretch}{1.4}
\caption{Exact inference prompt templates utilized for evaluating the models across three temporal grounding tasks. All paradigms receive the identical text prompt alongside the visual input to ensure fair comparison.}
\label{tab:prompts}
\resizebox{\textwidth}{!}{%
\begin{tabular}{@{}p{0.16\textwidth}p{0.82\textwidth}@{}}
\toprule
\textbf{Task (Dataset)} & \textbf{Inference Prompt Template} \\
\midrule
\textbf{Moment Retrieval} \newline (Charades-STA) &
\texttt{Localize the visual content described by the given textual query '\{query\}' in the video, and output the start and end timestamps in seconds.} \\
\midrule
\textbf{Highlight Detection} \newline (QVHighlights) &
\texttt{Please find the highlight contents in the video described by a sentence query, determining the highlight timestamps and its saliency score on a scale from 1 to 5. Now I will give you the sentence query: '\{query\}'. Please return the query-based highlight timestamps and salient scores.} \\
\midrule
\textbf{Dense Video Captioning} \newline (YouCook2) &
\texttt{Scrutinize the video and determine multiple occurrences, providing their initial and final timestamps as well as a summary of each action.} \\
\bottomrule
\end{tabular}%
}
\end{table*}

\noindent\textbf{Module Architectures and Loss.}
Table~\ref{tab:backbone_specs} summarizes the backbone-level specifications. Let $d$ denote the LLM hidden dimension. For the \textit{Continuous} paradigm, the Time Decoder is a 3-layer MLP with ReLU activations ($d \!\to\! d \!\to\! d \!\to\! 2(R{+}1)$, Xavier initialization) that maps the hidden state at the $\langle\text{TIME\_STAMP}\rangle$ position to $2{\times}(R{+}1)$ distribution logits over $R{=}32$ bins. A symmetric Time Encoder converts ground-truth timestamps into Gaussian distributions ($\sigma{=}1.0$) over $R{+}1$ anchor points and maps them back to $\mathbb{R}^{d}$ via an inverse MLP ($2(R{+}1) \!\to\! d \!\to\! d \!\to\! d$). The total loss is $\mathcal{L} = \mathcal{L}_{\mathrm{LM}} + \mathcal{L}_{\mathrm{DFL}} + \mathcal{L}_{\mathrm{DIoU}}$ with equal weights. For the \textit{Generative} paradigm, TimeTower and ScoreTower are each an embedding layer over a 13-token character vocabulary ($\texttt{Embedding}(13, d)$), and SyncTower is a single learnable embedding ($\texttt{Embedding}(1, d)$); all three heads use standard cross-entropy. The \textit{Text} paradigm adds no modules and uses the standard next-token loss. Additional hyperparameters: weight decay 0.01, LoRA dropout 0.05, seed 42, max sequence length 4096.

\noindent\textbf{Inference Post-Processing.}
For \textit{Continuous}, the $2(R{+}1)$ logits are split into start/end groups, softmax-normalized, and decoded via expectation $\hat{t} = \sum_{i=0}^{R} p_i \cdot i$, then rescaled by $\text{duration}/R$ to absolute seconds. For \textit{Generative}, the head-switching state machine accumulates character-level time tokens, splits by $\langle\text{sep}\rangle$, and parses floating-point timestamps; scores are decoded analogously. For \textit{Text}, timestamps are extracted from the generated string via pattern matching (e.g., ``from X to Y seconds'').

\noindent\textbf{Statistical Variance Analysis.}
\label{sec:variance}
To verify the robustness and reproducibility of our findings, we repeat training with 3 different random seeds on two representative backbones: SmolVLM2-2.2B (a compact model) and Molmo2-4B (a mid-scale model), covering the two primary scales studied in this work. Table~\ref{tab:variance} reports the mean and standard deviation of mIoU on Charades-STA and QVHighlights for all three paradigms.

\begin{table}[h]
\centering
\caption{Variance analysis across 3 random seeds on two representative backbones. We report mean$\pm$std of mIoU. The consistently small standard deviations confirm that the observed paradigm gaps are robust and not attributable to stochastic training variation.}
\label{tab:variance}
\small
\setlength{\tabcolsep}{3pt}
\begin{tabular}{ll cc}
\toprule
 & & Charades-STA & QVHighlights \\
Backbone & Para. & mIoU & mIoU \\
\midrule
\multirow{3}{*}{SmolVLM2-2.2B}
 & Text  & 20.6$\pm$0.4 & 16.3$\pm$0.3 \\
 & Cont. & 46.8$\pm$0.3 & 54.2$\pm$0.4 \\
 & Gen.  & 27.9$\pm$0.5 & 23.7$\pm$0.4 \\
\midrule
\multirow{3}{*}{Molmo2-4B}
 & Text  & 24.1$\pm$0.3 & 18.8$\pm$0.4 \\
 & Cont. & 56.1$\pm$0.2 & 59.8$\pm$0.3 \\
 & Gen.  & 32.5$\pm$0.4 & 27.3$\pm$0.3 \\
\bottomrule
\end{tabular}
\end{table}

\noindent As shown in Table~\ref{tab:variance}, the standard deviations across all configurations are consistently small (typically $\leq$0.5 mIoU), demonstrating that our conclusions are robust to random initialization. Notably, the performance gap between paradigms (e.g., \textit{Cont.} vs.\ \textit{Text}: $\sim$26 mIoU on SmolVLM2-2.2B) is an order of magnitude larger than the within-paradigm variance, confirming that the observed differences are attributable to the output formulation rather than stochastic training variation.

\section{Comprehensive Survey of VTG-MLLMs}
\label{app:survey_table}
To situate our empirical findings within the broader landscape of Video-LLM research, we compile a comprehensive performance leaderboard of existing state-of-the-art multimodal large language models capable of video temporal grounding. As presented in Table~\ref{tab:survey_mega}, we detail their zero-shot performance across standard benchmarks, strictly grouped by their underlying temporal output paradigms.

This macro-level overview highlights the extreme heterogeneity in current VTG-MLLM evaluation setups, strongly validating the necessity of our controlled, variable-isolating study presented in the main text.

\begin{table*}[h]
\centering
\small
\caption{Comprehensive performance leaderboard of existing VTG-MLLMs. Metrics represent zero-shot performance reported in their respective papers or official repositories. "-" indicates the absolute metric was not reported or evaluated by the original authors. Methods are grouped by their temporal output paradigm to facilitate direct comparison.}
\label{tab:survey_mega}
\resizebox{\textwidth}{!}{%
\begin{tabular}{@{}ll ccc cc ccc@{}}
\toprule
\multirow{3}{*}{\textbf{Method}} & \multirow{3}{*}{\textbf{Backbone}} & \multicolumn{3}{c}{\textbf{Charades-STA}} & \multicolumn{2}{c}{\textbf{QVHighlights}} & \multicolumn{3}{c}{\textbf{YouCook2}} \\
 & & \multicolumn{3}{c}{\textit{(Moment Retrieval)}} & \multicolumn{2}{c}{\textit{(Highlight Detection)}} & \multicolumn{3}{c}{\textit{(Dense Video Captioning)}} \\
\cmidrule(lr){3-5} \cmidrule(lr){6-7} \cmidrule(l){8-10}
 & & \textbf{R1@0.5} & \textbf{R1@0.7} & \textbf{mIoU} & \textbf{mAP} & \textbf{HIT@1} & \textbf{CIDEr} & \textbf{SODA$_c$} & \textbf{F1} \\
\midrule
\multicolumn{10}{l}{\textit{\textbf{Text Numeral Generation}}} \\
TimeChat~\cite{ren2024timechat} & LLaMA-2-7B & 32.2 & 13.4 & -- & 14.5 & 23.9 & 1.2 & 3.4 & 12.6 \\
VTimeLLM~\cite{huang2024vtimellm} & Vicuna-7B & 27.5 & 11.4 & 31.2 & -- & -- & 3.4 & 0.9 & -- \\
GroundingGPT~\cite{li2024groundinggpt} & Vicuna-7B & 29.6 & 11.9 & -- & -- & -- & -- & -- & -- \\
HawkEye~\cite{wang2024hawkeye} & Vicuna-7B & 31.4 & 14.5 & 33.7 & -- & -- & -- & -- & -- \\
Chrono-GPT~\cite{meinardus2024chrono} & GPT-4o & 28.8 & 11.0 & 33.0 & -- & -- & -- & -- & -- \\
\midrule
\multicolumn{10}{l}{\textit{\textbf{Temporal Token Generation}}} \\
SeViLA~\cite{yu2023self} & Flan-T5-XL (3B) & 15.0 & 5.8 & 18.3 & -- & -- & -- & -- & -- \\
Momentor~\cite{qian2024momentor} & LLaMA-7B & 26.6 & 11.6 & 28.5 & 7.6 & 17.0 & -- & -- & -- \\
VTG-LLM~\cite{guo2025vtg} & LLaMA-2-7B & 33.8 & 15.7 & -- & 16.5 & 33.5 & 5.0 & 1.5 & 17.5 \\
Grounded-VideoLLM~\cite{wang2024grounded} & Phi-3.5 (4B) & 36.4 & 19.7 & 36.8 & -- & -- & -- & -- & -- \\
TRACE~\cite{guo2024trace} & Mistral-7B & 40.3 & 19.4 & 38.7 & 42.7 & 26.8 & 8.1 & 2.2 & 22.4 \\
\midrule
\multicolumn{10}{l}{\textit{\textbf{Continuous Temporal Decoding}}} \\
TimeRefine~\cite{wang2026timerefine} & Vicuna-7B & 38.6 & 16.4 & 36.2 & -- & -- & -- & -- & -- \\
InternVideo2.5~\cite{wang2025internvideo2} & InternLM2.5-7B & 43.3 & 22.8 & 41.7 & 26.5 & 54.1 & -- & -- & -- \\
VideoMind~\cite{liu2025videomind} & Qwen2-VL-7B & 59.1 & 31.2 & 50.2 & -- & -- & -- & -- & -- \\
DisTime~\cite{zeng2025distime} & InternVL2.5-8B & 60.3 & 30.8 & 53.1 & -- & -- & 31.0 & 6.9 & 26.4 \\
\bottomrule
\end{tabular}%
}
\end{table*}

\section{Extended Failure Taxonomy and Qualitative Examples}
\label{app:qualitative}
To complement the failure case analysis in Section~\ref{sec:qualitative} of the main paper, we first provide a quantitative taxonomy of error types to systematically dissect paradigm weaknesses, followed by additional visual examples that concretely illustrate these behaviors.

\subsection{Quantitative Taxonomy of Errors}
We conduct a systematic diagnosis on the Charades-STA dataset to analyze the failure distribution across the three representative paradigms: DisTime (\textit{Continuous}), TRACE (\textit{Temporal Token}), and VtimeLLM (\textit{Text Numeral}). For each paradigm, we collect the failed predictions (i.e., IoU $< 0.5$) and categorize them into three distinct error types:
\begin{itemize}
    \item \textbf{Type A (Temporal Hallucination):} The model predicts a temporal window completely disjoint from the ground truth, often occurring when LLMs fail to ground abstract numerals to continuous video frames.
    \item \textbf{Type B (Boundary Jitter):} The model correctly identifies the macroscopic event, but the temporal boundaries are overly bloated or truncated, failing the strict 0.5 IoU threshold.
    \item \textbf{Type C (Semantic Failure):} The model fundamentally misunderstands the query (e.g., confusing "opening" with "closing") or the causal logic of the video.
\end{itemize}

\begin{figure}[h]
\centering
\includegraphics[width=0.9\linewidth]{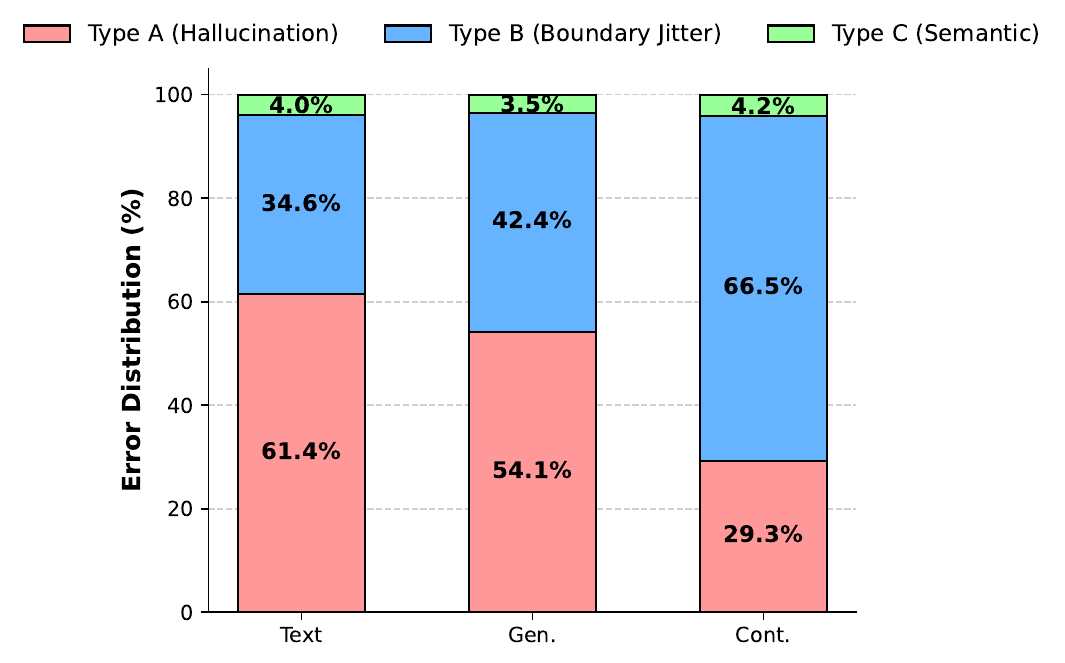}
\caption{Quantitative error taxonomy across representative paradigms on Charades-STA. The composition explicitly breaks down the failure cases (where IoU $< 0.5$) into three types. The \textit{Continuous} paradigm drastically shifts the error mode from severe hallucinations to minor boundary jitters.}
\label{fig:error_taxonomy}
\end{figure}

As visually detailed in Figure~\ref{fig:error_taxonomy}, the text numeral \textit{VtimeLLM} exhibits a staggering 61.4\% of its errors stemming from Temporal Hallucination (Type A). The token-based \textit{TRACE} also suffers heavily from hallucinations (54.1\%). Conversely, the continuous \textit{DisTime} model fundamentally shifts the error distribution: its hallucinations are drastically suppressed to 29.3\%, with the majority of its errors (66.5\%) being minor Boundary Jitters (Type B). This proves that regressing a continuous distribution prevents the LLM from arbitrarily "guessing" wild timestamps.

\subsection{Qualitative Visualizations}
\begin{figure*}[h]
\centering
\includegraphics[width=\textwidth]{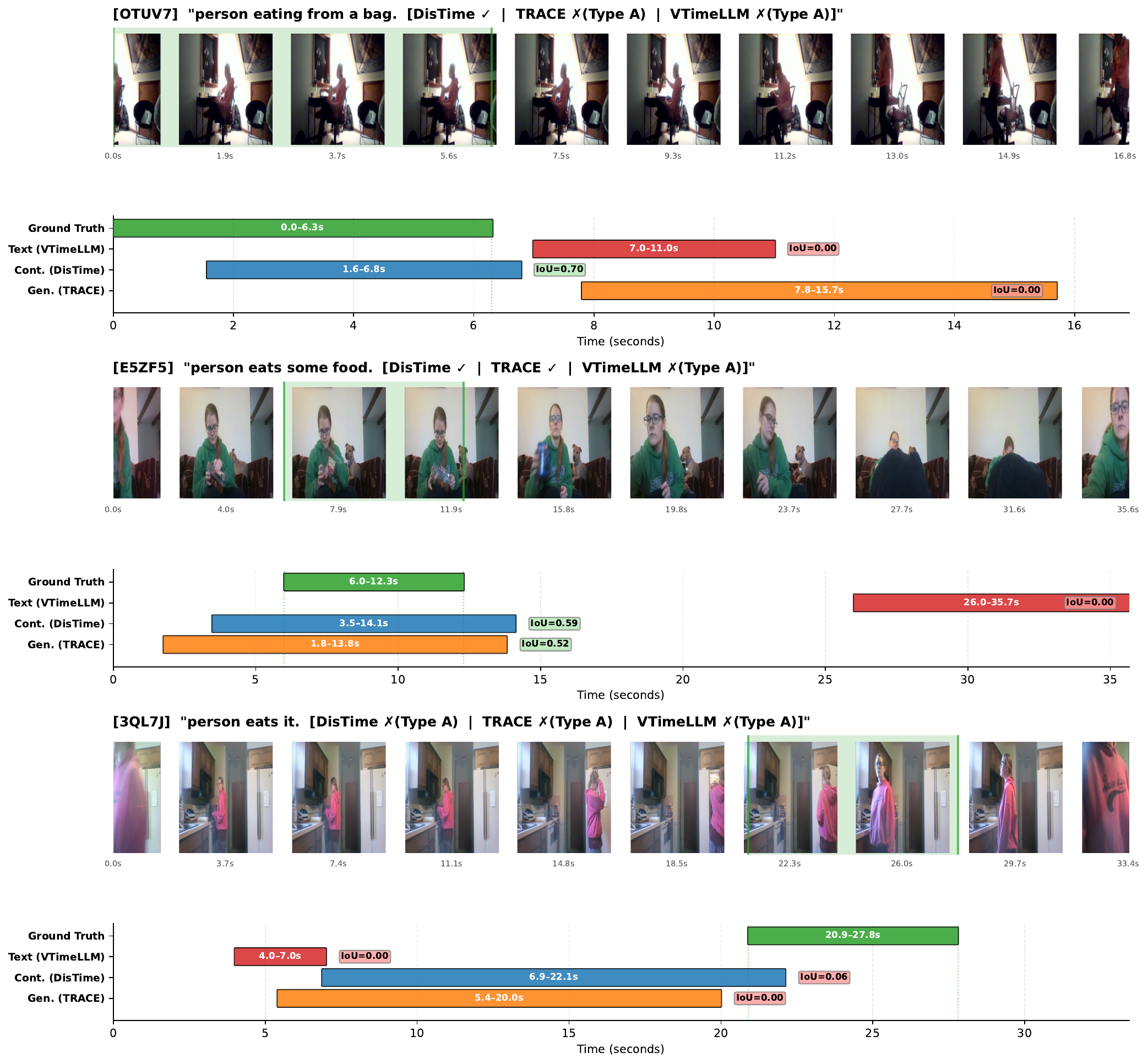}
\vspace{-6mm}
\caption{\textbf{Qualitative comparisons focusing on Type A (Temporal Hallucination) errors.} Green bars denote Ground Truth.
\textbf{(Top)} DisTime successfully localizes the event, whereas both discrete paradigms (TRACE and VTimeLLM) completely hallucinate early timeframes, demonstrating the continuous paradigm's robustness against blind guessing.
\textbf{(Middle)} DisTime and TRACE accurately capture the action, leaving VTimeLLM as the sole model exhibiting a severe Type A hallucination (predicting the very end of the video). This underscores the text-numeral paradigm's extreme susceptibility to temporal hallucinations.
\textbf{(Bottom)} A universally challenging query where all paradigms fail via hallucination, predicting disjoint early segments. This highlights that while continuous decoding mitigates hallucinations, extreme causal complexity can still trigger Type A errors across all architectures.}
\label{fig:qualitative_type_a}
\end{figure*}

\begin{figure*}[h]
\centering
\includegraphics[width=\textwidth]{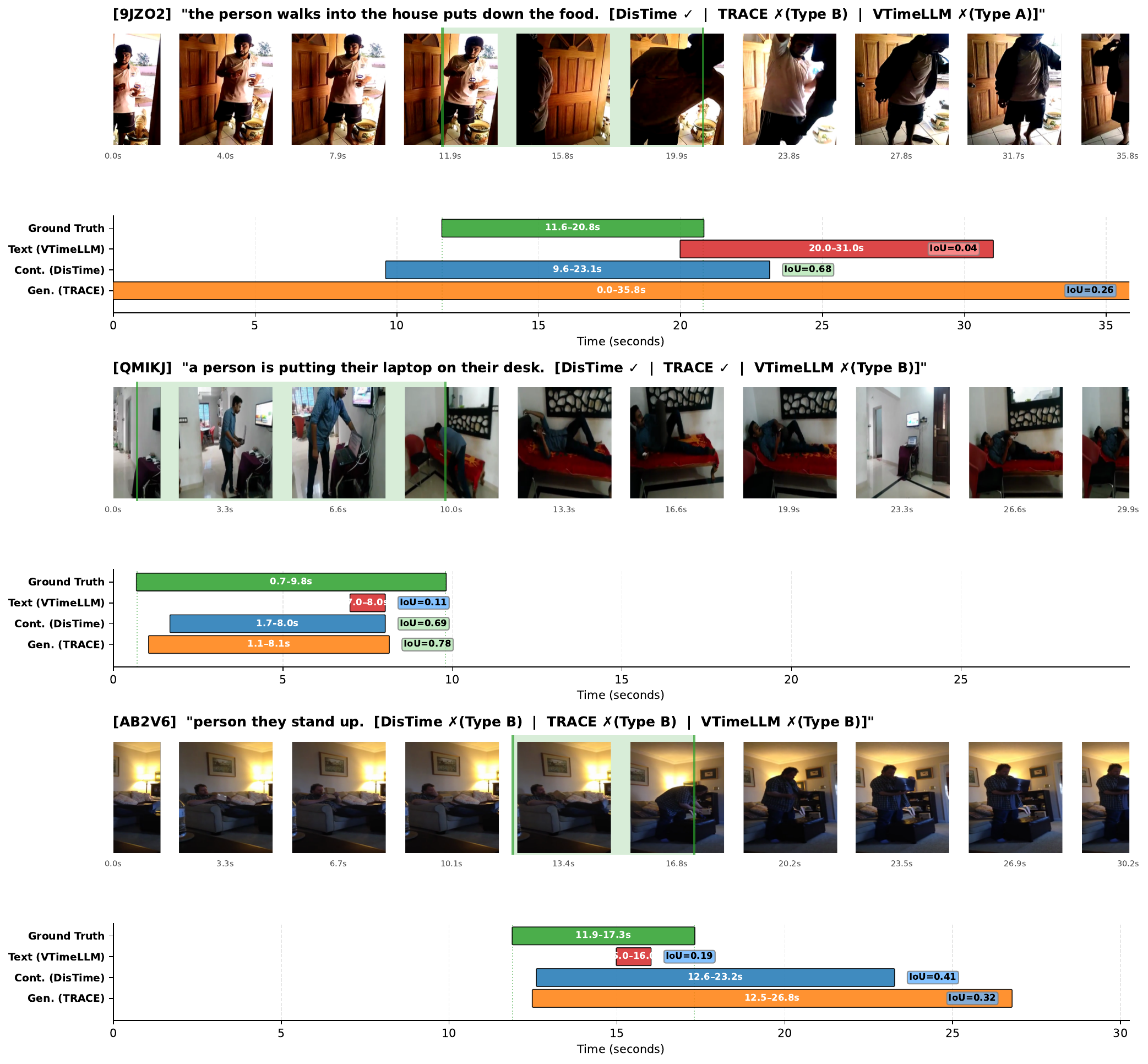}
\vspace{-6mm}
\caption{\textbf{Qualitative comparisons focusing on Type B (Boundary Jitter) errors.}
\textbf{(Top)} DisTime accurately covers the true temporal window. TRACE, however, exhibits massive ``Token Bloat'' (a severe Type B error spanning almost the entire video), while VTimeLLM hallucinates entirely (Type A).
\textbf{(Middle)} Both DisTime and TRACE successfully localize the action. VTimeLLM identifies the general vicinity but severely truncates the boundary (7.0-8.0s vs. GT 0.7-9.8s), yielding a failing IoU of 0.11.
\textbf{(Bottom)} A scenario characterized by ambiguous action boundaries where all models suffer from Type B errors. Notably, DisTime and TRACE regress overly bloated durations (over-extension), whereas the text-based VTimeLLM predicts an extremely narrow window (truncation), reflecting their distinct mechanisms for handling temporal uncertainty.}
\label{fig:qualitative_type_b}
\end{figure*}

\begin{figure*}[h]
\centering
\includegraphics[width=\textwidth]{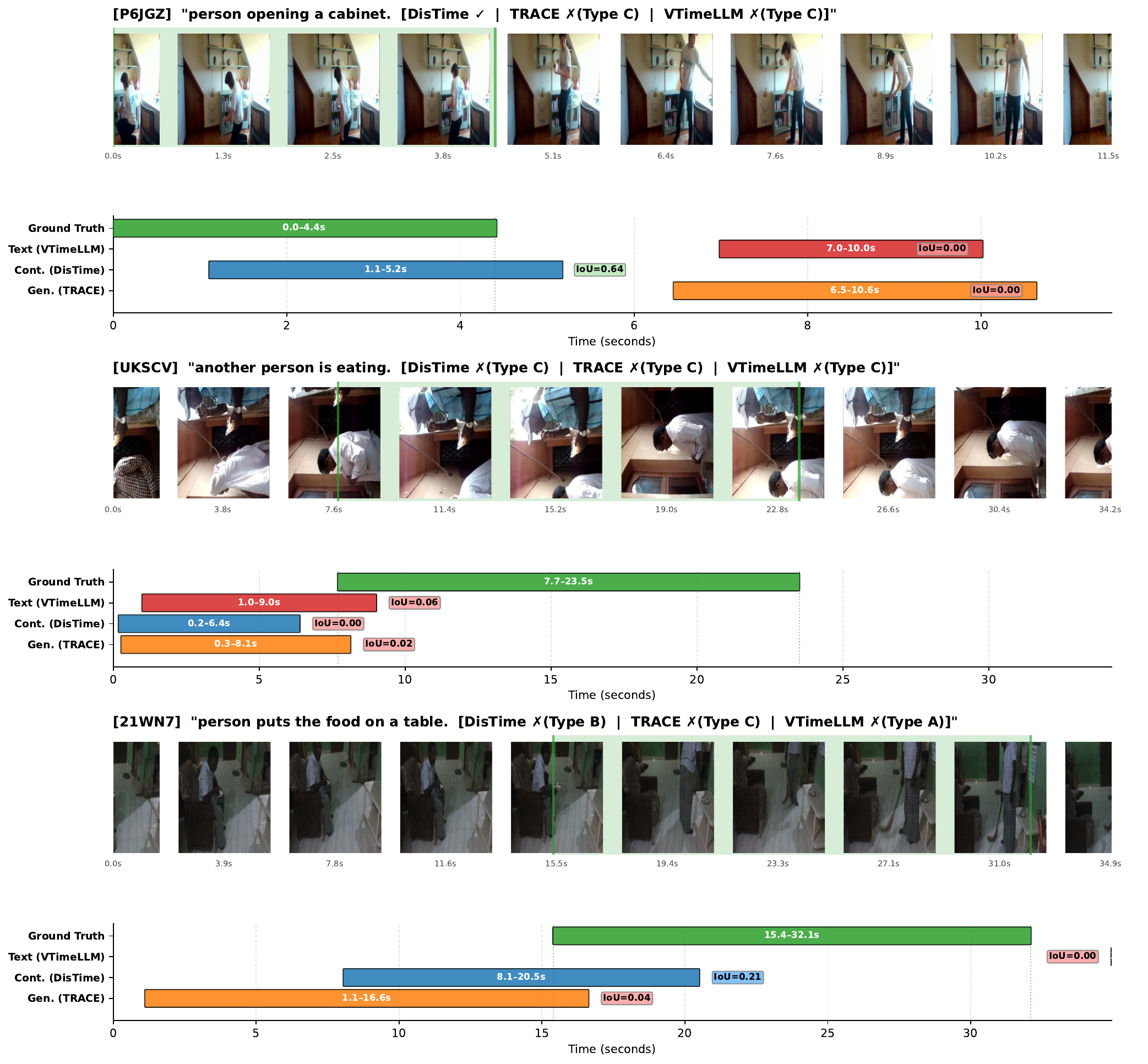}
\vspace{-6mm}
\caption{\textbf{Qualitative comparisons focusing on Type C (Semantic Confusion) errors.}
\textbf{(Top)} DisTime perfectly localizes ``opening a cabinet''. In contrast, both TRACE and VTimeLLM confuse this action with a semantically similar but chronologically later event, resulting in disjoint predictions (Type C).
\textbf{(Middle)} A highly deceptive case where all three paradigms are misled by earlier visual cues of a person eating, completely missing the ground-truth timeframe. This represents a universal semantic failure across current VTG-MLLMs.
\textbf{(Bottom)} A complex failure case demonstrating diverse error manifestations. DisTime successfully captures the event but fails due to benign boundary jitter (\textit{Type B}, IoU=0.21). Meanwhile, TRACE predicts a semantically incorrect early action (\textit{Type C}), and VTimeLLM hallucinates entirely off-target (\textit{Type A}). This perfectly visually encapsulates our taxonomy findings.}
\label{fig:qualitative_type_c}
\end{figure*}

Guided by the quantitative taxonomy above, we significantly expand our qualitative analysis by dedicating three comprehensive figures to concretely illustrate these distinct behavioral patterns. We systematically dissect the performance of three representative methods, across our proposed error types: Temporal Hallucination (Figure~\ref{fig:qualitative_type_a}), Boundary Jitter (Figure~\ref{fig:qualitative_type_b}), and Semantic Confusion (Figure~\ref{fig:qualitative_type_c}).

These exhaustive visual examples highlight a critical divergence in underlying failure modes. The \textbf{\textit{Continuous}} paradigm gracefully manages fuzzy action boundaries and naturally suppresses severe temporal hallucinations. As shown in the visualizations, even in challenging scenarios where DisTime fails the strict evaluation threshold (e.g., Figure~\ref{fig:qualitative_type_b} Bottom), its errors are predominantly benign \textit{Type B (Boundary Jitters)}, typically manifesting as an overly bloated temporal distribution that still directionally covers the actual event.

In stark contrast, constrained by the inherent mismatch between discrete token generation and the continuous temporal space, the discrete generative paradigms frequently struggle with strict timestamp regression. Both text-based and token-based approaches are highly susceptible to outputting confidently incorrect, completely disjoint timeframes (\textit{Type A}) or suffering from fundamental semantic misalignments (\textit{Type C}) when reasoning over complex videos.

\end{document}